\documentclass[letterpaper, 10 pt, conference]{ieeeconf}  

\IEEEoverridecommandlockouts                              

\usepackage{graphicx} 
\usepackage{float}
\usepackage[subrefformat=parens,labelformat=parens, caption = false]{subfig}
\usepackage{bm}
\usepackage[linesnumbered,ruled,vlined]{algorithm2e}
\usepackage{multirow}
\usepackage{makecell}
\SetKwInOut{Input}{Input}
\SetKwInOut{Output}{Output}
\usepackage{amsmath,amssymb,amsfonts}
\graphicspath{{./Figure/}}
\usepackage[sorting=none, style=ieee]{biblatex}
\usepackage[symbol]{footmisc}

\usepackage {tablefootnote}

\DeclareMathOperator*{\argmin}{argmin}
\title{\LARGE \bf
Motion Degeneracy in Self-supervised Learning of Elevation Angle Estimation for 2D Forward-Looking Sonar
}

\author{Yusheng Wang$^{1}$, Yonghoon Ji$^{2}$, Chujie Wu$^{1}$, Hiroshi Tsuchiya$^{3}$, Hajime Asama$^{1}$ and Atsushi Yamashita$^{4}$
\thanks{$^{1}$Y.~Wang, C.~Wu, H.~Asama are with the Department of Precision Engineering, Graduate School of Engineering, The University of Tokyo, Japan.~{\tt\small \{wang,chujie,asama\}@robot.t.u-tokyo.ac.jp}}
\thanks{$^{2}$Y. Ji is with the Graduate School for Advanced Science and Technology, JAIST, Japan. {\tt\small ji-y@jaist.ac.jp}}
\thanks{$^{3}$H. Tsuchiya is with the Research Institute, Wakachiku Construction Co., Ltd., Japan. {\tt\small hiroshi.tsuchiya@wakachiku.co.jp}}
\thanks{$^{4}$A.~Yamashita is with the Department of Human and Engineered Environmental Studies, Graduate School of Frontier Sciences, The University of Tokyo, Japan.~{\tt\small yamashita@robot.t.u-tokyo.ac.jp}}
}

\begin{document}

\maketitle
\thispagestyle{empty}
\pagestyle{empty}

\begin{abstract}

2D forward-looking sonar is a crucial sensor for underwater robotic perception. A well-known problem in this field is estimating missing information in the elevation direction during sonar imaging. There are demands to estimate 3D information per image for 3D mapping and robot navigation during fly-through missions. Recent learning-based methods have demonstrated their strengths, but there are still drawbacks. Supervised learning methods have achieved high-quality results but may require further efforts to acquire 3D ground-truth labels. The existing self-supervised method requires pretraining using synthetic images with 3D supervision. 
This study aims to realize stable self-supervised learning of elevation angle estimation without pretraining using synthetic images. Failures during self-supervised learning may be caused by motion degeneracy problems. We first analyze the motion field of 2D forward-looking sonar, which is related to the main supervision signal. We utilize a modern learning framework and prove that if the training dataset is built with effective motions, the network can be trained in a self-supervised manner without the knowledge of synthetic data. Both simulation and real experiments validate the proposed method.

\end{abstract}

\section{INTRODUCTION}
Sonar sensors play a vital role in underwater perception as ultrasound is invariant to water turbidity and illumination conditions. The next generation of 2D forward-looking sonar (FLS) can generate images with high resolution and good quality. These sensors are also small in size and are suitable for use in commercial-class remotely operated vehicles (ROVs) and autonomous underwater vehicles (AUVs). Furthermore, they have already been applied to underwater robotics tasks, such as mosaicking, mapping, and navigation \cite{Hurtos2015,Wang2020JOE,Negahdaripour2013}. Owing to the unique imaging principle, the elevation angle information is missing during image formation. Therefore, one of the most focused but challenging topics in FLS research is the retrieval of 3D information.  

Early research focused on the sparse 3D reconstruction of acoustic images. Point and line features were used to represent the 3D model, which is unintuitive for human comprehension \cite{Mai20171}. Recently, researchers have focused on dense 3D reconstruction based on acoustic images. Generally, these methods can be classified into multiple- and single-view methods. Multi-view methods use numerous images to generate 3D models. They usually require a large number of viewpoints, making it necessary to hover around the target using underwater vehicles \cite{guerneve2018,Qadri2023} or rotate the camera with motors \cite{Wang2020JOE}. Single-view methods utilize shadow information or model ultrasound propagation for 3D reconstructions \cite{Aykin2016}. Such methods only work under ideal conditions and are not general and robust. Recent deep learning-based methods have shown promising results for the single-view missing dimension estimation problem \cite{DeBortoli2019,Wangicra2021}. The network in \cite{Wangicra2021} was trained in a supervised manner, which required 3D supervision. Because acquiring ground-truth information for underwater scenes can be difficult due to the limitations of sensing techniques, a self-supervised training method would be highly valuable. Although in \cite{DeBortoli2019}, the missing dimension estimation network can be learned in a self-supervised manner, it requires pretraining using synthetic images of a similar target with ground truth labels, which limits the application of the method because it may require 3D model of a similar scene, and the preparation of synthetic data also requires considerable effort. Self-supervised learning of elevation angle estimation for FLS without the help of simulation remains an open problem. 

\begin{figure}[tb]
\centering
{\includegraphics[width=1.0\columnwidth]{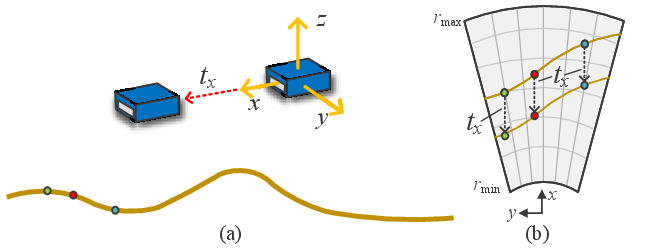}}
\caption{An example of motion degeneration. (a) The 2D forward-looking sonar moves along the $x$-axis with a small motion $t_x$. (b) All the 2D pixels in the acoustic image move $-t_x$ in the $x$-axis. The 2D pixel displacement is independent of the 3D structure. Such data may deteriorate the self-supervised learning process of elevation angle estimation.} 
\label{fig:xmotion}
\end{figure}

The aim of this study was to solve the aforementioned problem. Self-supervised learning of elevation angle estimation uses two adjacent frames from the acoustic video, namely, the target view image $I_t$ and the source $I_s$. A synthetic target view image $\Tilde{I_t}$ can be generated from $I_s$, motion $M_{t\rightarrow s}$, and estimated pixel-wise target view elevation angle $E_t=g(I_t)$. Here $g(.)$ denotes the neural network.
 $\Tilde{I_t}$ can be expressed as
$\tilde{I_t} = \pi(I_s,E_t,M_{t\rightarrow s})$,
where $\pi(.)$ denotes the inverse warping process.
The basic idea is to optimize the following equation:
 \begin{equation}
     \argmin_{E_t}\mathcal{L}(I_t, \Tilde{I_t}),
 \end{equation}
 where $\mathcal{L}$ is the loss function to minimize the difference between the original target image and the one from view synthesis.

We find that the motion type during dataset collection is crucial, which may lead to training failure if the motion is improper. As shown in Fig.~\ref{fig:xmotion}, for some basic motions such as $x$-axis translation, the displacements of the pixels are independent of the 3D structure. In other words, an arbitrary $E_t$ may satisfy Eq.~(1), which degrades the training of network $g$. To determine the type of motion that may contribute to training, we first analyze the motion field of the acoustic images. The motion field provides the relationship between pixel displacement, pixel position, sensor motion, and elevation angle. Based on theoretical analysis, we utilize the most efficient motions to generate the dataset. We then build a learning framework using modern techniques. For real noisy images, we use a pre-trained network to generate a signal mask to filter the background noise and multi-path reflections for better performance.    

In summary, our contributions are as follows.
\begin{itemize}
    \item We propose a self-supervised learning method for FLS missing dimension estimation. Furthermore, we prove that the knowledge of synthetic data is not essential.
    \item We analyze the motion field for acoustic images and choose the effective motions to train the missing dimension estimation network.
    \item Extensive experiments including simulations and field experiments were carried out to verify our method. 
\end{itemize}

\section{Related Works}
\subsection{Acoustic Camera 3D Reconstruction}
Early research on FLS used point or line features to retrieve the 3D information of an object. Several methods have been proposed to estimate the 3D position of the features and the pose of the camera simultaneously \cite{Mai20171,Huang2015,Huang2016}. However, such sparse representation is not intuitive for human perception. Feature extraction and data association are also problems for acoustic images, considering the low signal-to-noise ratio (SNR). Wang et al. applied a Gaussian process with AKAZE features for terrain reconstruction \cite{wangj2019}. Recent works have also used two acoustic cameras to sense the environment \cite{Negahdaripour2018b,McConnell2020}. However, they still require the extracted features to achieve stable performance.

Dense 3D reconstruction has been a trend in recent studies. One direct approach is to use multiple views to construct a 3D model. Akyin et al. applied space carving to small objects using signal masks \cite{Aykin2017}. Guerneve et al. linearized the sonar projection model to an orthogonal projection and applied min filtering to achieve a carving scheme \cite{guerneve2018}. Wang et al. used occupancy mapping with an inverse sensor model to probabilistically carve the space and update an object \cite{Wang2020JOE}. Westman et al. utilized non-line-of-sight for 3D reconstruction \cite{Westman2020icra,Westman2020iros}. Qadri et al. proposed an implicit neural representation for surface reconstruction \cite{Qadri2023}. The aforementioned methods achieve convincing results but require numerous viewpoints. Although in \cite{Wang2022}, attempts have been made to estimate 3D information with 2$\sim$3 viewpoints, it requires training with ground truth labels, which is challenging in underwater environments.  

There is a need to estimate 3D information with a single image for 3D mapping and navigation through a fly-through motion. The acoustic camera can be considered as a confocal configuration of a spotlight and camera. Shape-from-shading scheme can be applied to such problems. Aykin et al. modeled the ultrasound propagation and assumed diffuse reflection for underwater objects \cite{Aykin2016}. 3D models can be generated from object contours. Westman et al. utilized a similar scheme for 3D reconstruction of flat continuous surfaces \cite{Westman2019iros}. The above-mentioned methods have strong assumptions regarding the scene and may fail owing to the low SNR of the acoustic image. 

Deep learning techniques have been applied to estimate 3D information from a single image. Wang et al. proposed a method for estimating the front depth of an acoustic image. It can solve the non-bijective correspondence problem \cite{Wangicra2021}. However, the network was trained in a supervised manner, where ground truth labels were required. Although labels can be acquired by generative adversarial network (GAN) and a simulator, it is necessary to model a similar scene in the simulator \cite{liu2021}. Considering the difficulty of acquiring ground truth labels, self-supervised learning methods have also been proposed. DeBortoli et al. proposed a learning-based method for elevation angle estimation from a single acoustic image \cite{DeBortoli2019}. The network was first trained in a supervised manner using synthetic images with ground-truth labels and then fine-tuned using self-supervised signals for real images. However, it is unclear whether the network can be trained without pretraining using synthetic images. In fact, during training, supervised learning and self-supervised learning were alternately carried out to ensure performance. It is also necessary for deeper discussions of the problem, such as which motion contributes to the training process, because FLS faces a severe motion degeneracy problem. In this study, we analyze the motion field of the FLS and choose efficient motions to build the dataset for self-supervised training. Using a state-of-the-art training framework and proper datasets, we train the network with acoustic video and corresponding motion information only.

\subsection{Monocular Camera Self-supervised Depth Estimation}

With the rapid development of deep learning technologies, depth estimation from a single optical image has shown stable performance. Initially, most of the methods were fully supervised, requiring ground truth depth during training \cite{Eigen2014,Liu2016}. However, this is challenging in real-world settings, considering the difficulty of acquiring ground truth labels. Garg et al. proposed a method for training a depth estimation network in a self-supervised manner \cite{Garg2016}. A pair of images, the source, and the target with a known small camera motion were used to train the network. 
Zhou et al. also used view synthesis as the supervision signal but proved that camera motion can also be learned simultaneously \cite{Zhou2017}. 
Pure rotation is a well-known degenerate motion in perspective cameras. The transformation of the two image frames becomes homography, where 3D information is no longer necessary. Considering the projection principle of perspective cameras, pure rotations may not contribute to depth estimation training \cite{Van2019}. Improving the performance on indoor datasets with more rotational motion remains an open problem. Some recent studies have begun to consider pure rotation degeneracy \cite{Bian2022}. The FLS faces a severe but unique motion degeneracy problem that will be addressed and discussed in this study. We focus here on the estimation of the elevation angle for a given motion and leave the motion estimation to future work.   
\section{Problem Analysis}
\subsection{Inverse Warping Signal}
The core of self-supervised learning is to determine the optimal $E_t$ based on Eq.~(1) and use a network $g(.)$ to learn it. The motion between the two frames must be small to guarantee sufficient overlap and photometric constancy. As mentioned above, there are cases in which $E_t$ cannot be solved owing to degenerate motions. To analyze the degenerate motions, we notice that the self-supervised learning approaches are highly related to the concept of the motion field. The motion field models pixel displacement when the sensor has a small movement. Denoting the pixel displacement as $\frac{ds}{dt}$ and the motion field estimation process as $f(.)$, motion field estimation can be described as follows.
\begin{equation}
\frac{d\bm{s}}{dt} = f(\bm{s}, E_t, M_{t\rightarrow s}),
\end{equation}
where $\bm{s}$ denotes the position of the 2D pixel.
Using $\frac{d\bm{s}}{dt}$ and $I_s$, it is possible to generate the synthetic target view $\tilde{I_t}$. This study uses the motion field model to investigate the self-supervised learning of the elevation angle estimation problem.
\subsection{Motion Model}
\begin{figure}[tb]
\centering
{\includegraphics[width=1.0\columnwidth]{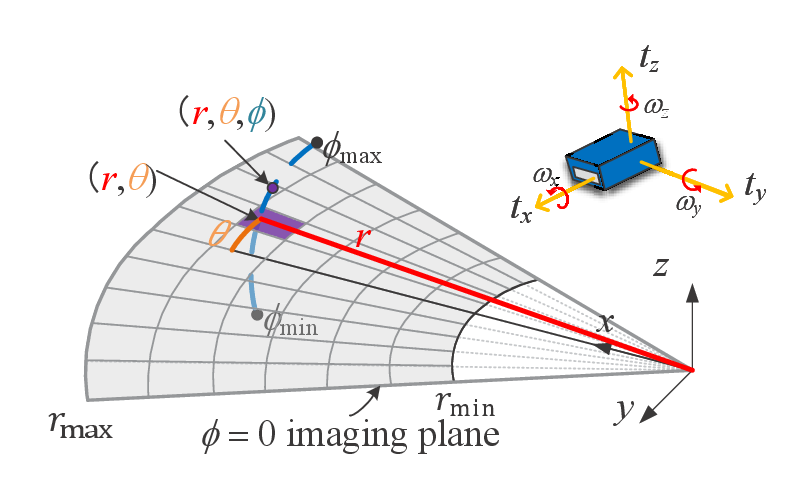}}
\caption{Projection model and the definition of motion. It can be equivalently seen as a projection of a 3D point to the $\phi=0$ plane, where the elevation information is missing. Small translations along $x$, $y$ and $z$ axes are defined as $t_x$, $t_y$ and $t_z$, respectively. Small rotations along $x$, $y$ and $z$ axes are defined as $\omega_x$, $\omega_y$ and $\omega_z$.} 
\label{fig:projection}
\end{figure}

First, we explain the 2D forward-looking sonar projection model, as shown in Fig.~\ref{fig:projection}. A 3D point in the sonar coordinate system is written as follows.

\begin{equation}
\bm{p_s} = 
\begin{bmatrix}
r\cos \phi \cos \theta & r \cos \phi \sin \theta & r \sin \phi 
\end{bmatrix}^{\top}.
\end{equation}

The position of the 2D pixel $\bm{s}$ is as follows.
\begin{equation}
\bm{s} = \begin{bmatrix}
x_s & y_s 
\end{bmatrix}^{\top}
=
\begin{bmatrix}
r \cos \theta & r  \sin \theta  
\end{bmatrix}^{\top}.
\end{equation}

Thus, the relationship between $\bm{p_c}$ and $\bm{s}$ can be determined.
\begin{equation}
\bm{s}
=
\frac{1}{\cos{\phi}}
\begin{bmatrix}
1 & 0 & 0\\
0 & 1 & 0
\end{bmatrix}
\bm{p_s}
.  
\end{equation}

The displacement $\frac{d\bm{s}}{dt}$ is calculated by differentiating $\bm{s}$ with respect to $dt$ based on Eq.~(5).
\begin{equation}
\frac{d\bm{s}}{dt} = \frac{1}{\cos{\phi}}\begin{bmatrix}
1 & 0 & 0\\
0 & 1 & 0
\end{bmatrix}
\frac{d\bm{p_s}}{dt}+\tan{\phi}\bm{s}\frac{d\phi}{dt},
\end{equation}
where $\frac{d\phi}{dt}$ can be further represented by $\frac{d\bm{p_s}}{dt}$ as follows.
\begin{equation}
\frac{d\phi}{dt} = \frac{1}{r}\begin{bmatrix}
-\cos{\theta}\sin{\phi} & -\sin{\theta}\sin{\phi} & \cos{\phi}
\end{bmatrix}\frac{d\bm{p_s}}{dt}.
\end{equation}

When there is a small motion $(\bm{t},\bm{\omega})$ of the FLS, the displacement of the stationary 3D point in FLS coordinates follows rigid body movement which can be written as follows.
\begin{equation}
\frac{d\bm{p_s}}{dt} = -\bm{\omega}\times\bm{p_s}-\bm{t}.
\end{equation}

After combining Eq.~(6) to Eq.~(8), the following relationship between the displacement and sonar motion can be acquired, which is also known as the motion field.
\begin{equation}
\begin{bmatrix}
\frac{dx_s}{dt} \\\\ \frac{dy_s}{dt}
\end{bmatrix}=
\begin{bmatrix}
-\frac{t_x}{\cos{\phi}}-(\frac{t_z\sin{\phi}}{r})x_s-\omega_z y_s\\+(\frac{\sin{\phi}\tan{\phi}t_x}{r^2})x_s^2+(\frac{\sin{\phi}\tan{\phi}t_y}{r^2}\\+\frac{\tan{\phi}\omega_x}{r})x_sy_s+(\frac{\tan{\phi}\omega_y}{r})y_s^2
 \\
\\
-\frac{t_y}{\cos{\phi}}-(\frac{t_z\sin{\phi}}{r})y_s+\omega_z x_s\\+(\frac{\sin{\phi}\tan{\phi}t_y}{r^2})y_s^2+(\frac{\sin{\phi}\tan{\phi}t_x}{r^2}\\-\frac{\tan{\phi}\omega_y}{r})x_sy_s-(\frac{\tan{\phi}\omega_x}{r})x_s^2
\end{bmatrix}.
\end{equation}

Here, we conduct a discussion based on ARIS EXPLORER 3000, where the aperture angle in the elevation angle ranges from [-7$^\circ$,7$^\circ$]. Then, $\cos{\phi}$ ranges from [0.9925,1], which is reasonable to approximate to 1. In addition, $\sin{\phi}$ ranges from [-0.1219,0.1219] and $\tan{\phi}$ ranges from [-0.1228,0.1228], such that $\sin{\phi}\tan{\phi}$ ranges from [0,0.015], and the second-order term is approximated to zero. The first-order terms, $\sin{\phi}$ and $\tan{\phi}$ are maintained. Then, Eq.~(9) becomes
\begin{equation}
\begin{bmatrix}
\frac{dx_s}{dt} \\\\ \frac{dy_s}{dt} 
\end{bmatrix}=
\begin{bmatrix}
-{t_x}-(\frac{t_z\sin{\phi}}{r})x_s-\omega_z y_s\\+(\frac{\tan{\phi}\omega_x}{r})x_sy_s+(\frac{\tan{\phi}\omega_y}{r})y_s^2
 \\\\
 -{t_y}-(\frac{t_z\sin{\phi}}{r})y_s+\omega_z x_s\\-(\frac{\tan{\phi}\omega_y}{r})x_sy_s-(\frac{\tan{\phi}\omega_x}{r})x_s^2
\end{bmatrix}.
\end{equation}

In \cite{Negahdaripour2013}, motion fields were used for motion estimation of FLS according to a flat surface. Although an equation similar to Eq.~(9) was induced, it was assumed that the second-order terms of the pixel position could be ignored. In other words, the terms with $x_s^2$ and $y_s^2$ were ignored. This may not hold in many cases because the units of $x_s$ and $y_s$ are meters. In fact, especially for $x_s$, it may range from several meters to tens of meters, and the second-order may be a huge value.

\subsection{Basic Motion Influence}
In this subsection, the influence of the basic motion on the motion field is discussed. Specifically, the following motions are discussed:1) $x$-axis, $y$-axis translations, and $z$-axis rotation; 2) $x$-axis rotation; 3) $y$-axis rotation;  and 4)~$z$-axis translation.

\subsubsection{$x$-axis, $y$-axis translations and $z$-axis rotation}
By setting the other motion parameters to zero, the displacement caused by horizontal motion (i.e., $x$-axis, $y$-axis translations and $z$-axis rotation) can be written as follows. 
\begin{equation}
\begin{bmatrix}
\frac{dx_s}{dt} \\\\ \frac{dy_s}{dt} 
\end{bmatrix}=
\begin{bmatrix}
 -{t_x}-\omega_z y_s \\\\ -{t_y}+\omega_z x_s
\end{bmatrix}.
\end{equation}

Apparently, the displacement is independent of the elevation angle $\phi$. In other words, the elevation angle could not be retrieved from the aforementioned motion. Notably, horizontal motion is frequently used for underwater robot navigation, if the imaging plane is parallel to the motion plane, 3D information cannot be acquired.
\subsubsection{$x$-axis rotation}
$x$-axis rotation is a well-known effective motion for 3D reconstruction FLS. Empirically, a small $x$-axis rotation can generate large displacement. This study provides a mathematical explanation with the following equation. 
\begin{equation}
\begin{bmatrix}
\frac{dx_s}{dt} \\\\ \frac{dy_s}{dt} 
\end{bmatrix}\! =\!
\begin{bmatrix}
(\frac{\tan{\phi}\omega_x}{r})x_sy_s \\\\ -(\frac{\tan{\phi}\omega_x}{r})x_s^2
\end{bmatrix}
\! =\!
\begin{bmatrix}
r\omega_x{\tan{\phi}}\cos{\theta}\sin{\theta}\\\\-r\omega_x{\tan{\phi}}{\cos^2{\theta}}
\end{bmatrix}.
\end{equation}

As a typical case, where $r=3.5$~m, $\phi=3.5^\circ$, and $\omega_x=10^\circ$ when the azimuth angle changes from $-15^\circ$ to $15^\circ$, the displacement is shown in Fig.~\ref{fig:sensitive}(a). In this equation, $\cos^2{\theta}$ is relatively large, which makes the displacement obvious. In this figure, $\rho$ refers to the range resolution equal to 0.003~m and $\gamma$ is the tangential resolution when $r=3.5$~m. For $\frac{dx_s}{dt}$, a change smaller than $\rho$ cannot be detected in the acoustic image. For $\frac{dy_s}{dt}$, a change smaller than $\gamma$ is difficult to detect. This may require subpixel optimization for displacement estimation. Notably, the ARIS EXPLORER 3000 has one of the best resolutions for FLS. The resolution range for Blueview P900 is approximately 0.0254 m, which is not sufficiently sensitive to detect some of the displacements for a close range. However, sensors such as Blueview are designed to be applied to a larger scene, and the discussions in this section also stand by scaling up the problem with a larger range and coarser resolution. 

\subsubsection{$y$-axis rotation}

For $y$-axis rotation, the displacement is dependent on the elevation angle $\phi$ in the following equation. Theoretically, $\phi$ can be obtained from the $y$-axis rotation. However, in practice, $y$-axis rotation is not very efficient  \cite{Wang2020JOE}. 

\begin{equation}
\begin{bmatrix}
\frac{dx_s}{dt} \\\\ \frac{dy_s}{dt} 
\end{bmatrix}\!\!\!=\!\!\!
\begin{bmatrix}
(\frac{\tan{\phi}\omega_y}{r})y_s^2 \\\\ -(\frac{\tan{\phi}\omega_y}{r})x_sy_s
\end{bmatrix}\!\!\!=\!\!\!
\begin{bmatrix}
 r\omega_y{\tan{\phi}}{\sin^2{\theta}}\\\\ -r\omega_y{\tan{\phi}}\cos{\theta}\sin{\theta}
\end{bmatrix}\!.
\end{equation}

With the same configuration ($r=3.5$~m, $\phi=3.5^\circ$, $\omega_y=10^\circ$), Figure~\ref{fig:sensitive}(b) shows the sensitivity analysis of $y$-axis rotation. From the results, it is known that the displacement is quite small, which may make it difficult to detect and influence the 3D reconstruction performance.

\subsubsection{$z$-axis translation}

$z$-axis translation is also considered an effective motion that has been used for 3D reconstruction. 

\begin{equation}
\begin{bmatrix}
\frac{dx_s}{dt} \\\\ \frac{dy_s}{dt} 
\end{bmatrix}=
\begin{bmatrix}
 -(\frac{t_z\sin{\phi}}{r})x_s \\\\ -(\frac{t_z\sin{\phi}}{r})y_s
\end{bmatrix}=
\begin{bmatrix}
-{t_z\sin{\phi}}\cos{\theta} \\\\ -{t_z\sin{\phi}}\sin{\theta}
\end{bmatrix}.
\end{equation}

\begin{figure*}[bt] 
	\begin{center}
		\centering
		\subfloat[\label{roll}]{\includegraphics[width=0.65\columnwidth]{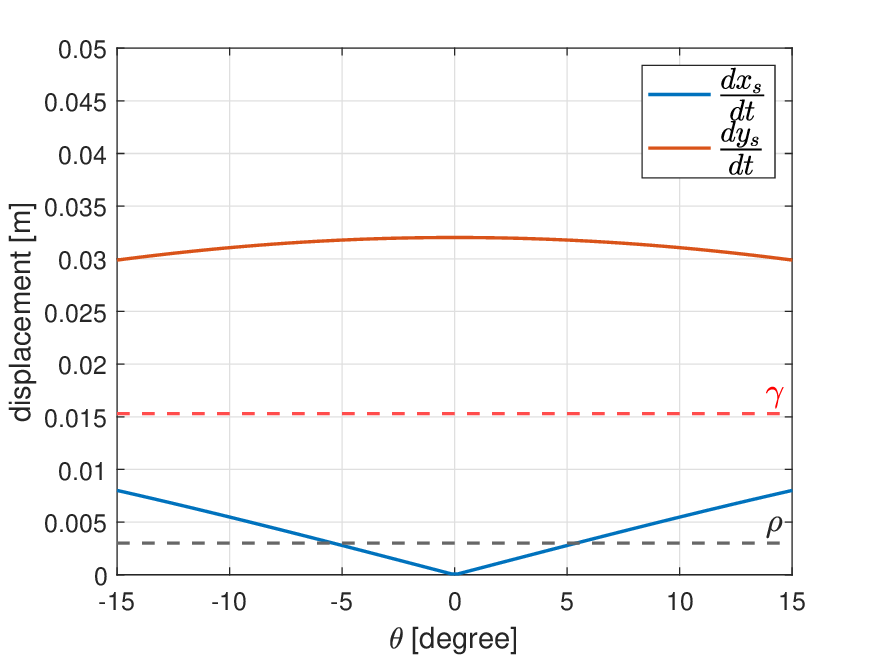}}\enskip 
		\subfloat[ \label{pitch}]{\includegraphics[width=0.65\columnwidth]{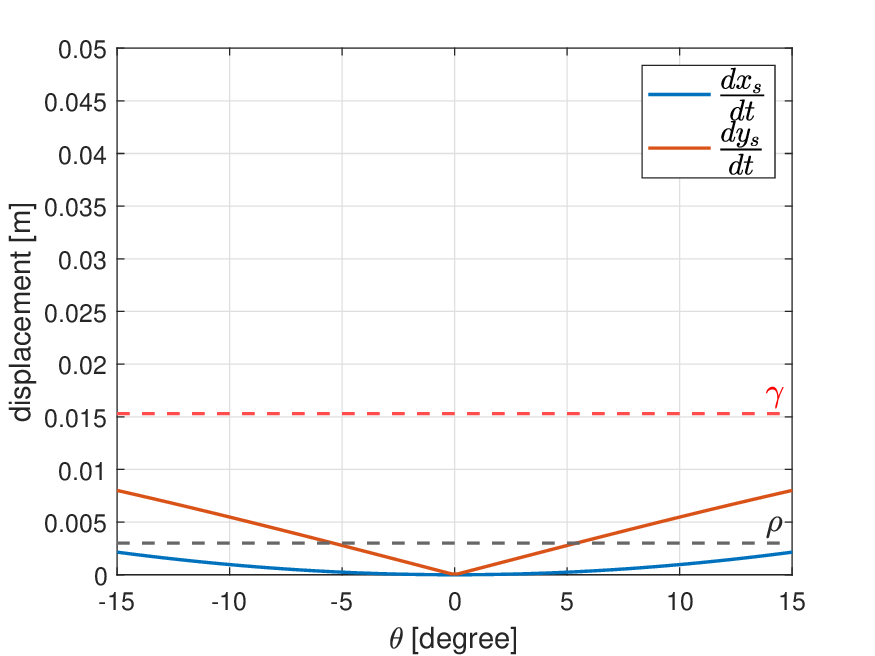}}\enskip
		\subfloat[ \label{z}]{\includegraphics[width=0.65\columnwidth]{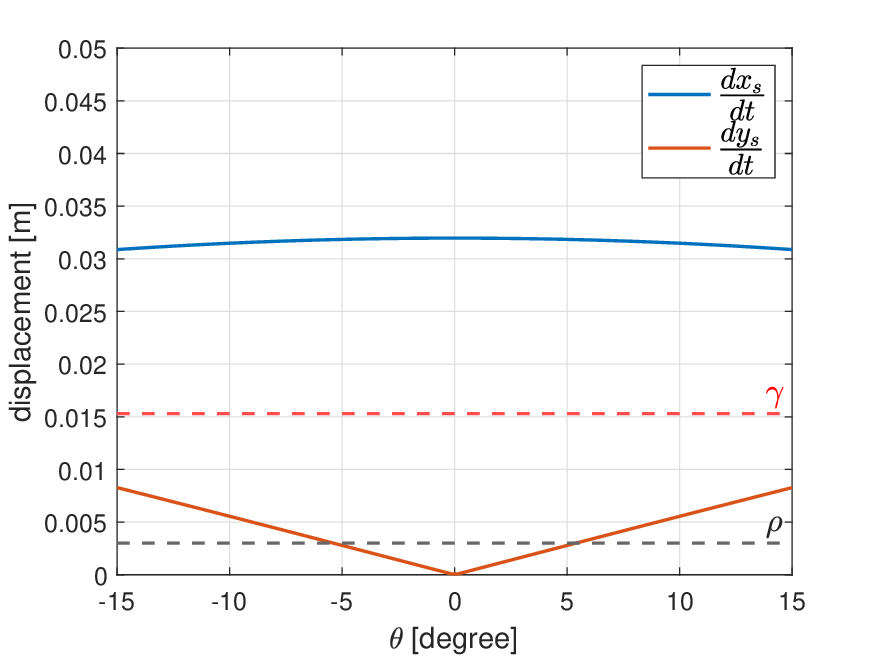}}
		\caption[Sensitivity analysis]{Sensitivity analysis. (a) The displacement caused by $\omega_x$, where $r=3.5$~m, $\phi=3.5^\circ$, and $\omega_x=10^\circ$. (b) The displacement caused by $\omega_y$, where $r=3.5$~m, $\phi=3.5^\circ$, and $\omega_y=10^\circ$. (c)~The displacement caused by $t_z$, where $r=3.5$~m, $\phi=3.5^\circ$, and $t_z=0.1745$~m.}  
		\label{fig:sensitive}
	\end{center}
\end{figure*}

For the sensitivity analysis, with a configuration of $r=3.5$~m, $\phi=3.5^\circ$, and $t_z=0.1745$~m, the results are shown in Fig. ~\ref{fig:sensitive}(c). The value of $\cos{\theta}$ in Eq.~(14) is quite large, which makes the $z$-axis translation an effective motion. 

The following conclusions can be drawn.
\begin{itemize}
	\item Basic motions $x$-axis rotation and $z$-axis translation are sensitive to $\phi$, which are effective for tasks like 3D reconstruction.
	\item The horizontal motions $t_x,t_y,\omega_z$ are independent to $\phi$ in motion field, which do not contribute to 3D reconstruction.
	\item The displacement from $\omega_y$ is small, which may be difficult to be detected. 
\end{itemize}

For dataset generation, basic motions $x$-axis rotation and $z$-axis translation are theoretically effective.

\section{End-to-end Learning}
\subsection{Learning Framework}
\begin{figure}[tb]
\centering
{\includegraphics[width=1.0\columnwidth]{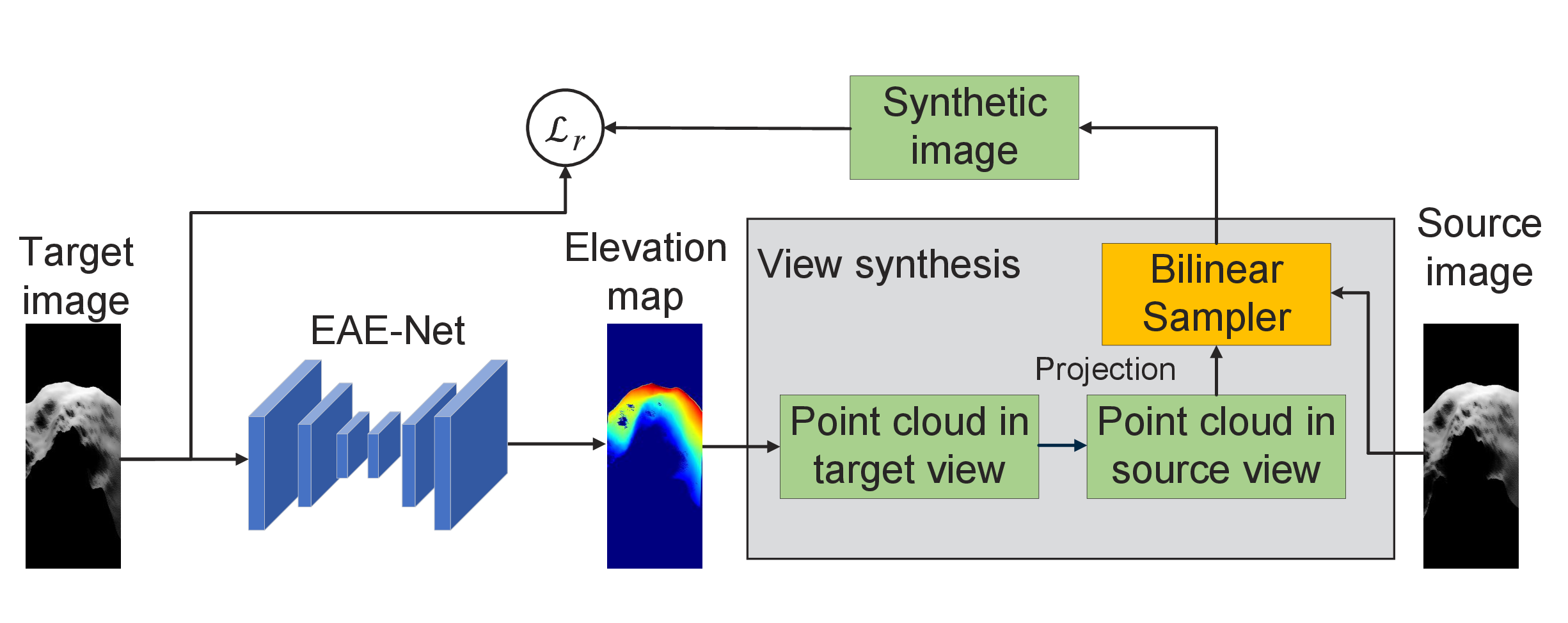}}
\caption{The illustration of the self-supervised learning framework during training. For test time, it is possible to estimate the elevation map per image using EAE-Net. 
} 
\label{fig:eaenet}
\end{figure}

The learning framework is illustrated in Fig.~\ref{fig:eaenet}. We name the network for \underline{e}levation \underline{a}ngle \underline{e}stimation as EAE-Net. We use a UNet-structure with a Sigmoid layer before the final output. The elements in the elevation map are 0$\sim$1. Denoting the aperture angle in the elevation direction as $\phi_{\rm aperture}$, the values in the elevation maps are linearly mapped to [$-\frac{\phi_{\rm aperture}}{2}$,$\frac{\phi_{\rm aperture}}{2}$] for further processing. By specifying the output range, the training process has better convergence. Because there is no 3D supervision, two adjacent frames from the acoustic video are used for training. We first estimate the target view elevation map $E_t$ and transfer it to the point cloud. With known motion $M_{t\rightarrow s}$, the point cloud is transferred to the source coordinates. By sampling the source image $I_s$, it is possible to generate the synthetic target image $\tilde{I_t}$. Bilinear sampling is used to ensure the process is differentiable. The network can be optimized by minimizing the difference between $I_t$ and $\tilde{I_t}$. Implementation of inverse warping $\pi(.)$ is expressed as Algorithm~\ref{algorithm1}.

\begin{algorithm}[h]
	\DontPrintSemicolon 
	\Input{Elevation map in target view $E_t$, source image $I_s$, motion $M_{t\rightarrow s}$ }
	\Output{Synthetic target view image from source image $\tilde{I_t}$}
	\ForEach {$(r_t,\theta_t, \phi_t) \in E_t $}
	{Transform the point to the Euclidean coordinates $\bm{p_t}$ using Eq.~(3).\\
		$\bm{p_s}= M_{t\rightarrow s}\bm{p_t}$.\;
		Project $\bm{p_s}$ to the image coordinates by generating $(r_s,\theta_s)$ pairs.\;
		Calculate the intensity $i_s$ of each point using bilinear sampling. \;
		$\tilde{I_t}$($r_t,\theta_t$) = $i_s$.\;
	}
	\Return $\tilde{I_t}$\;
	\caption[View synthesis in FLS]{View synthesis in FLS}
	\label{algorithm1}
\end{algorithm}

\subsection{Loss Functions}
To compare the synthetic target image $\tilde{I_t}$ and the real target image $I_t$, L1 and SSIM loss are used. Together they are called the reconstruction loss $\mathcal{L}_r$. 
\begin{equation}
\mathcal{L}_r = \beta\mathcal{M}_t(1-{\rm SSIM}(I_t,\tilde{I_t}))+(1-\beta)\mathcal{M}_t||I_t-\tilde{I_t}||_1,
\end{equation}
where $\beta$ is a hyperparameter that balances the two losses, and $\mathcal{M}_t$ is the mask of the target image. We set $\beta$ as 0.3.  

Smoothness loss $\mathcal{L}_s$ is implemented on the elevation maps in the target view ${E_t}$. The edge-ware smoothness loss is expressed as follows. It is assumed that the regions between the sharp edges are smooth. This is to smooth the elevation map when there is not much texture in the target view. 
\begin{equation}
\mathcal{L}_s = |{\partial r}{\mathcal{M}_tE_t}|e^{-|{\partial r}I_t|}+|{\partial \theta}{\mathcal{M}_tE_t}|e^{-|{\partial \theta}I_t|}.
\end{equation}

The total loss can be expressed as follows.
\begin{equation}
\mathcal{L} = \lambda_r\mathcal{L}_r+\lambda_s\mathcal{L}_s,
\end{equation}
where $\lambda_r$ and $\lambda_s$ are weights to balance the losses. We set $\lambda_r$ to 2 and $\lambda_s$ to 1 here.
\subsection{Signal Mask}
For real images, the background noise and multi-path reflections may influence the training process. This study focuses more on the geometric process during self-supervised learning and the influence of noise should be eliminated. For most 3D reconstruction works, the signal mask is essential and acquired by methods like binarization and region detection \cite{Wang2020JOE,Westman2019iros}. This study uses a direct but efficient method. We manually label the region with informative signals, and the remaining pixels are considered to be noise. By creating a small dataset with real image and signal mask pairs, it is possible to train a network for binary segmentation. We use a UNet structure as the network and binary cross entropy (BCE) with Dice loss to train the network \cite{Jadon2020}.

\section{Experiment}
\subsection{Simulation Experiment}
\subsubsection{Dataset Generation}
We used our simulator in Blender to build the synthetic dataset\footnote{https://github.com/sollynoay/Sonar-simulator-blender}. Here, we synthesized scenes of seabed terrain using A.N.T. Landscape add-on in Blender using hetero noise. Because the elevation angle estimation problem potentially suffers from the non-bijective 2D-3D correspondence problem \cite{Wangicra2021}, one pixel may correspond to multiple ground truth elevation angle values. To avoid this, in the synthetic dataset, most of the pixels in the acoustic image correspond to only one 3D position for a better evaluation. We generated two different terrains for training, three different terrains for validation, and three different terrains for test. We generated training datasets with six basic motions and evaluate them on one test dataset. The configurations of the datasets are listed in Table~\ref{tab:Table_dataset}. During training, consecutive three frames were used: the current frame was the target frame, and the past and the future frames were the source frames. For the motion between two adjacent frames, each motion follows uniform distribution and the scope is shown in Table~\ref{tab:Table_dataset}. Here, $\omega_y$ motion is smaller compared to the other rotational motions because with a larger $\omega_y$ value, the overlap of the consecutive frames may be insufficient. 
\begin{table}[ht]
	\centering
	\caption{Basic motion datasets}\label{tab:Table_dataset}
	\scalebox{1.0}{
		\begin{tabular}{|c|c|c|c|c|}
			\hline
			Dataset & Training (triplet) & Training (Motion)&Val&Test\\ 
                \hline
                $t_x$ & 3,000 & 8$\sim$12 [cm]& \multirow{6}{*}{1,500} & \multirow{6}{*}{1,500}\\
                \cline{1-3}
                $t_y$ & 3,000 & 8$\sim$12 [cm]& &\\
                \cline{1-3}
                $t_z$ & 3,000 & 8$\sim$12 [cm]& &\\
                \cline{1-3}
                $\omega_x$ & 3,000 & 5$\sim$10 [deg]& &\\
                \cline{1-3}
                $\omega_y$ & 3,000 & 2$\sim$4 [deg] & &\\
                \cline{1-3}
                $\omega_y$ & 3,000 & 5$\sim$10 [deg]& &\\
                \cline{1-3}
			\hline
		\end{tabular}
  }
\end{table}
\subsubsection{Implementation}
This method was implemented using PyTorch. The network was trained and tested using an NVIDIA GeForce RTX 3090. We set the learning rate to 0.0005 and batch size to 4. We used the Adam optimizer and trained the network for 15 epochs for each dataset.
\subsubsection{Metrics}
To evaluate the results, we used the mean absolute error (MAE) of the elevation map. We also transferred the elevation map into point cloud and used the chamfer distance (CD) for the evaluation. The ground truth point cloud is from the entire sensible region in the case of the non-bijective 2D-3D correspondence problem \cite{Wangicra2021}. 

\begin{equation}
    {\rm MAE} = \frac{\beta}{HW} \times \sum_{i=1}^{H}\sum_{j=1}^{W}|\hat{D}(i,j)-D(i,j)|,
\end{equation}
\begin{equation}
    {\rm CD} = \frac{\nu}{S_1}\sum_{x \in S_1} \min_{y \in S_2}||x-y||_2^2+\frac{\nu}{S_2}\sum_{y \in S_2} \min_{x \in S_1}||x-y||_2^2,
\end{equation}
where $\beta$ and $\nu$ are scale parameters, which refer to 1,000 and 500, respectively. $S_1$ and $S_2$ refer to the two point sets. We also used a percentage metric that measures the overall performance of accuracy and completeness with an error distance smaller than a threshold as the $f$-score \cite{Knapitsch2017}. For MAE and CD, the results are the smaller the better. The opposite is true for $f$-score. 
\subsubsection{Results}

\begin{figure}[tb]
\centering
{\includegraphics[width=1.0\columnwidth]{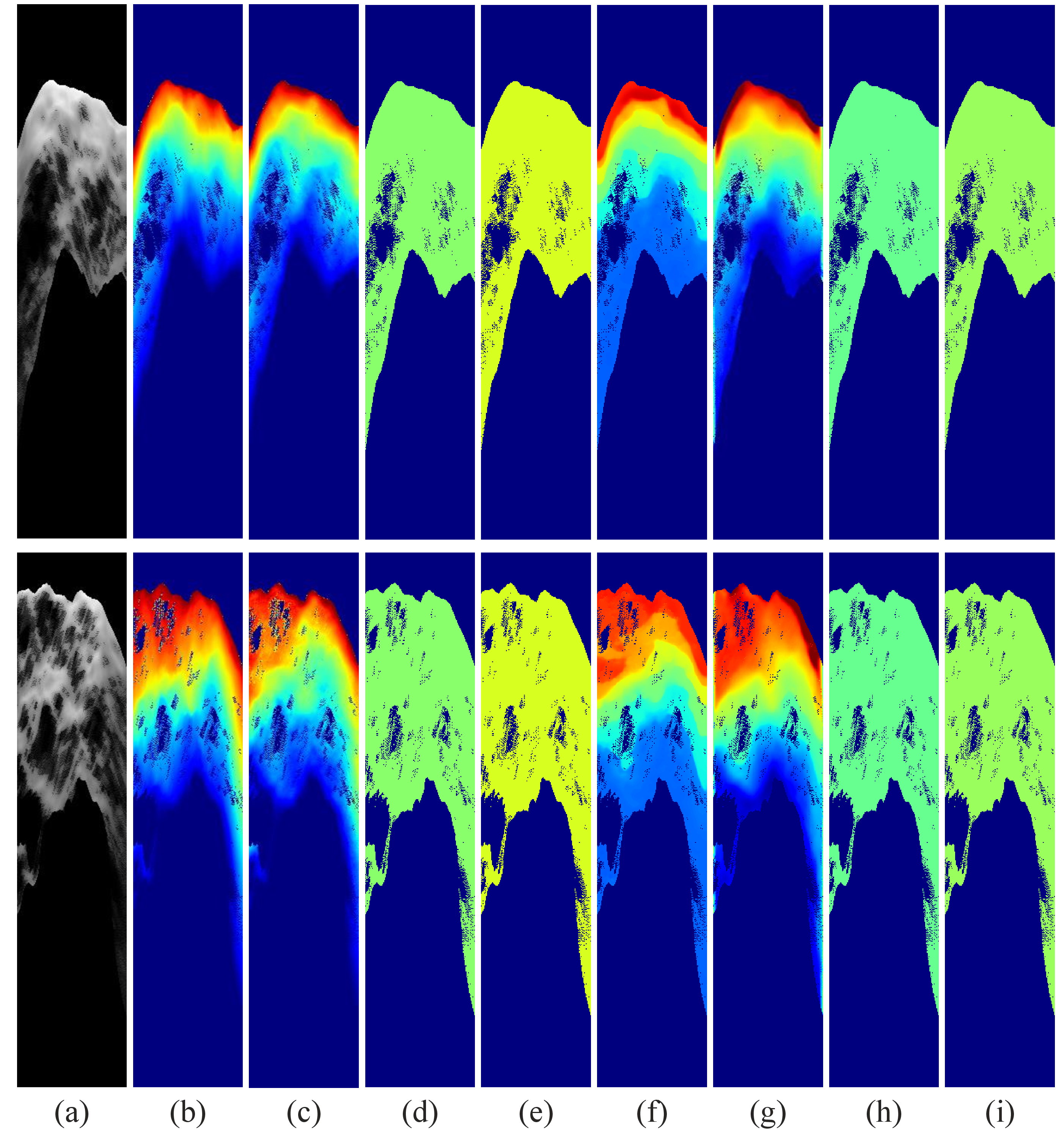}}
\caption{Visualization examples of the elevation angle estimation results. Color refers to the elevation angle. Images from left to right: (a) input image, (b) ground truth, (c) supervised learning results, (d) results from $t_x$ dataset, (e) results from $t_y$ dataset, (f) results from $t_z$ dataset, (g) results from $\omega_x$ dataset, (h) results from $\omega_y$ dataset, and (i) results from $\omega_z$ dataset. For the self-supervised method, only $\omega_x$ and $t_z$ datasets can successfully train the network.} 
\label{fig:result}
\end{figure}

\begin{table}[ht]
	\centering
	\caption{Simulation experiment results}\label{tab:Table_eae}
	\scalebox{0.95}{
		\begin{tabular}{|c|c|c|c|c|c|c|}
			\hline
		 & & EA & \multicolumn{3}{|c|}{Point cloud}	\\
            \hline
			& \makecell{3D \\ label} & MAE [rad]  & CD [m] & \makecell{$f$-score \\ ($<$1 mm)}  & \makecell{$f$-score \\ ($<$3 mm)}  \\ 
			\hline
                EAE & $\circ$  & 0.0234  & 1.210 & 72.71 &  90.61  \\
                \hline
                A2FNet& $\circ$ & --  & 1.784 & 63.52 & 84.77   \\
			\hline
                \hline
				$t_x$& $\times$ &  0.0956 & 161.300 & 0.06 &  0.18  \\
			\hline
                $t_y$& $\times$& 0.1086  & 191.566 & 0.02 &  0.08  \\
			\hline
                $t_z$& $\times$ & 0.0361 & 3.437 & 50.29 &  73.95  \\
			\hline
                $\omega_x$& $\times$ & \textbf{0.0298} & \textbf{1.972} & \textbf{59.81} &  \textbf{83.26}  \\
			\hline
                $\omega_y$& $\times$ & 0.1086 & 164.500 & 0.00 &  0.00  \\
			\hline
                $\omega_z$&  $\times$ & 0.0981  & 176.321 & 0.00 &  0.00  \\
			\hline
		\end{tabular}
	}
\end{table}

The quantitative results are shown in Fig.~\ref{fig:result}. EA refers to elevation angle. We also compared the proposed method with supervised learning methods. EAE refers to elevation angle estimation using supervised learning. A2FNet estimates the front depth map instead of the elevation map~\cite{Wangicra2021}. The signal mask in the simulation experiment is generated by binarization with a threshold. It can be known that $t_z$ and $\omega_x$ datasets can help the network learn the elevation angle estimation process. On the other hand, with other datasets, the network can barely learn any information. This proves the conclusion in Section~III. $t_x$, $t_y$, and $\omega_z$ datasets perform poorly due to the motion degeneracy. For $\omega_y$ datasets, the results are poor due to the sensitivity of the sensor. It is unsurprising that supervised learning with ground truth can generate more accurate results. However, by using $\omega_x$ dataset, the network can achieve the same level as A2FNet. This indicates that with proper training, self-supervised learning methods can approach supervised learning. One of the keys is using efficient motion. Since in terrain datasets, most of the pixels only correspond to one 3D position, direct elevation angle estimation may outperform A2FNet. 

Figure~\ref{fig:result} shows examples of elevation angle estimation results. It is quite obvious that only $t_z$ and $\omega_x$ datasets can contribute to training. For other motions, the network may converge to a random value. This is reasonable because based on our motion field analysis, an arbitrary elevation map may satisfy Eq.~(1), so that the network may perform randomly. After iterations and with the help of smoothness loss, the output of the network becomes a random constant. It can be also known that $\omega_x$ motion is more effective. The results from the $t_z$ dataset may contain some artifacts. This may be because the $\omega_x$ motion cause more overlaps and the photometric constancy holds better.

\subsection{Real Experiment}
\subsubsection{Dataset}
\begin{figure}[bt] 
	\begin{center}
		\centering
		\subfloat[\label{exp1}]{\includegraphics[width=0.475\columnwidth]{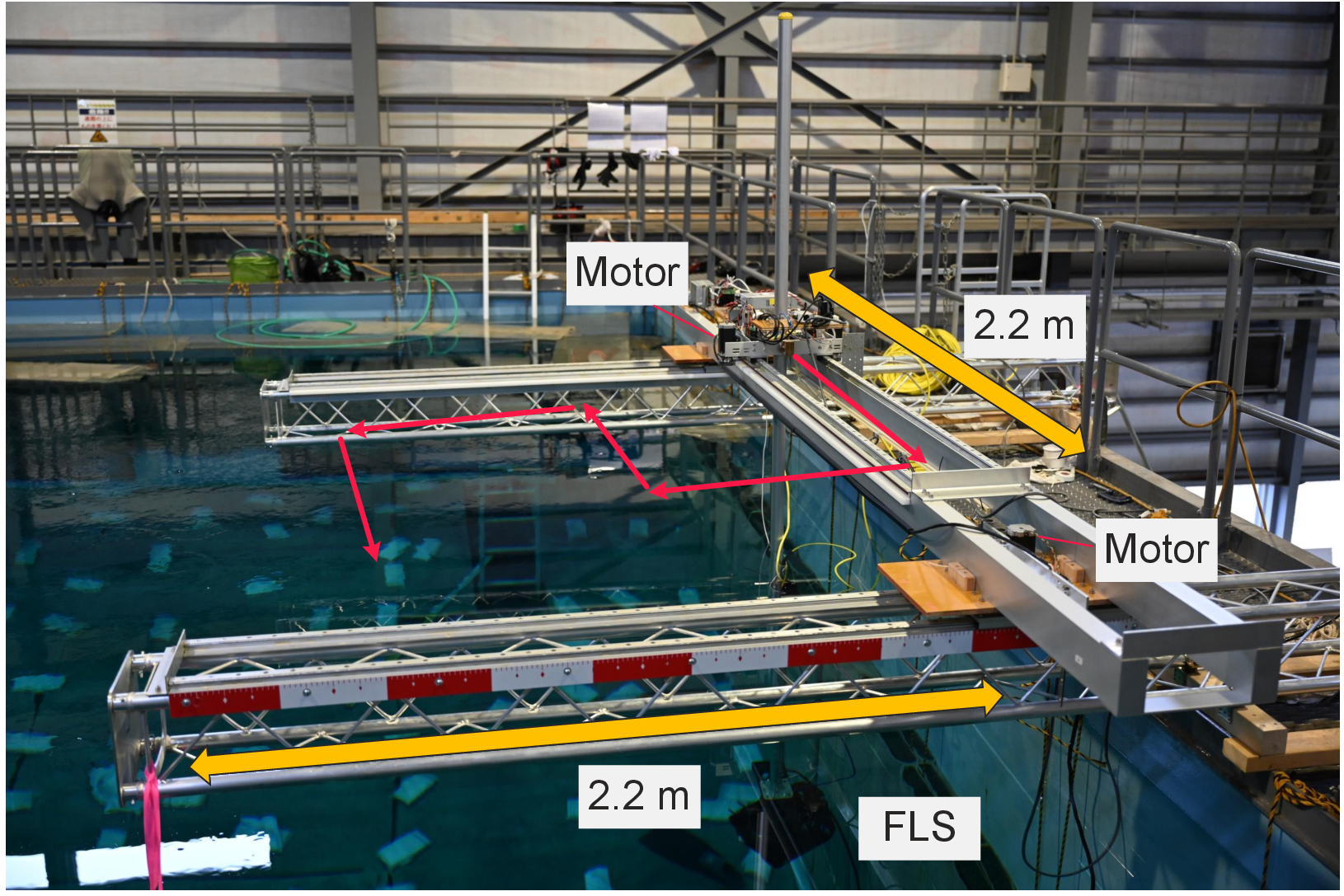}}\enskip 
		\subfloat[ \label{exp2}]{\includegraphics[width=0.475\columnwidth]{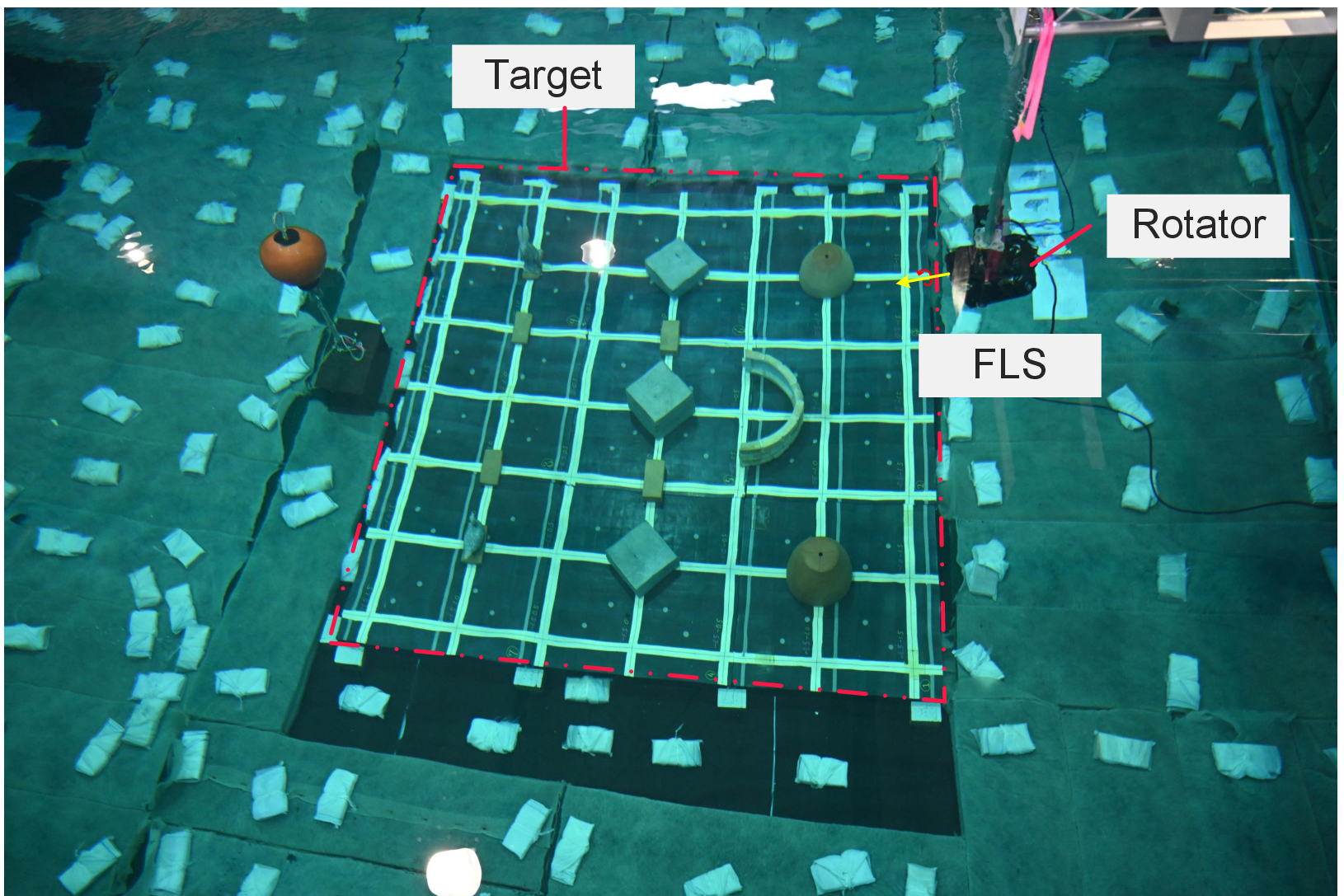}}
		\caption[Real experiment]{Real experiment: (a) moving device, and (b) target scene.}  
		\label{fig:watertank}
	\end{center}
\end{figure}

\begin{figure}[t] 
	\begin{center}
		\centering
            \subfloat[\label{raw}]{\includegraphics[width=0.475\columnwidth]{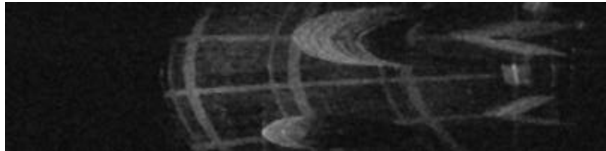}}\enskip 
		\subfloat[ \label{filter}]{\includegraphics[width=0.475\columnwidth]{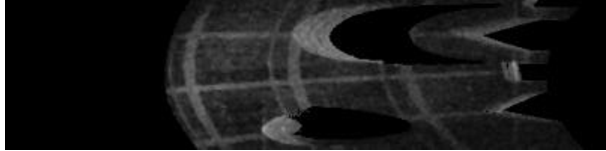}}
		\caption{Examples of the acoustic images in the dataset. (a) shows the raw image where there is noise in the non-signal region. Multi-path reflections may lie in the shadow region which may influence the learning process. (b)~is the image after filtering by the signal mask. }  
		\label{fig:img}
	\end{center}
\end{figure}

We built an $\omega_x$ dataset in a large-scale water tank to evaluate the proposed method. We used ARIS EXPLORER 3000 under 3.0 MHz mode. As shown in Fig.~\ref{fig:watertank}(a), the FLS was moved by a moving device. The FLS was carried to 51 positions. For each position, AR2 rotator was used to generate $\omega_x$ motion. The FLS rotates along $x$-axis from -35$^{\circ}$ to 35$^{\circ}$. We used images from 33 positions to train the network, 9 positions for validation, and the other 9 positions for test. In total, we used 1,071 images for training, 283 images for validation, and 294 images for test. We also used triplets to train the network. Denoting the roll angle for the target frame as $\alpha^{\circ}$, the roll angles for the two source frames were approximately $(\alpha-10)^{\circ}$ and $(\alpha+10)^{\circ}$, respectively. The target scene is illustrated in Fig.~\ref{fig:watertank}(b). The 3D ground truth of objects in the scene was measured in the land environment \cite{Wang2022}. They were set up in the water tank with a known configuration so it is possible to acquire the CAD model of the scene. It is also necessary to know the relative pose between the scene and the FLS. The initial value is measured by a diver using a ruler and the accurate result is obtained by comparing the synthetic image with the real image in the edge domain \cite{wang2022oceans}. After acquiring one pose accurately, the rest can be calculated with the help of control input for the moving device and the rotator. To prepare the signal masks, we manually labeled 164 images in the training set and trained the mask segmentation network. We set the batch size to 4 and the learning rate to 0.0003. The network was then trained for 25 epochs. Then, for all the acoustic images in the dataset, we multiplied the binary mask by the acoustic image to filter the background noise and multi-path reflection signals. As shown in Fig.~\ref{fig:img}, Figure~\ref{fig:img}(a) is an example of a raw acoustic image. After filtering the noise using the mask, the image in Fig.~\ref{fig:img}(b) was used for further processing. To train the EAE-Net, we used the same parameter setting as in the simulation experiment. 

\subsubsection{Results}
The quantitative results of the real experiment are listed in Table~\ref{tab:Table_real}. 
\begin{table}[ht]
	\centering
	\caption{Real experiment results}\label{tab:Table_real}
	\scalebox{0.95}{
		\begin{tabular}{|c|c|c|c|c|c|c|}
			\hline
		 &  & EA & \multicolumn{3}{|c|}{Point cloud}	\\
            \hline
		& \makecell{3D \\ label}	 & MAE\tablefootnote{Ground truth elevation angle for some pixels is not unique.} [rad]  & CD [m] & \makecell{$f$-score \\ ($<$1 mm)}  & \makecell{$f$-score \\ ($<$3 mm)}  \\ 
			\hline
                EAE & $\circ$ & 0.0139  & 0.8450 & 83.43 &  94.43  \\
                \hline
                A2FNet &$\circ$ & --  & 0.5385  & 92.67  &  97.68    \\
			\hline
                \hline
				$\omega_x$ w/o SM & $\times$ &  0.0364 & 5.243 & 43.16 &  66.44  \\
                \hline
                $\omega_x$ w SM & $\times$ & \textbf{0.0246} & \textbf{2.729}  &  \textbf{57.42} &  \textbf{83.18}  \\
            \hline
		\end{tabular}
	}
\end{table}
SM refers to the learned signal mask. For the self-supervised method, we also discussed the influence of the signal mask. It is known that using an $\omega_x$ dataset, the network can be successfully trained without pretraining on the synthetic images. Although not as good as supervised learning results, self-supervised learning with a learned signal mask can generate results with high quality. The $f$-score with a 3~mm threshold is over 83\%, indicating that most of the points have an error smaller than 3~mm. Considering the resolution in the range direction is 3~mm, this is acceptable for 3D reconstruction with a single image. If we directly generate the mask by binarization with a threshold, the mask is noisy and the multi-path signals cannot be filtered. The result may become slightly worse than that with a learned mask. A2FNet here exhibits better performance because, in the real dataset, there are more cases where one pixel corresponds to multiple 3D positions. The MAE metric here is only for reference because there are multiple ground truth elevation maps. For point cloud evaluation, the ground truth point cloud is the entire sensible region \cite{Wangicra2021}.

\begin{figure}[t] 
	\begin{center}
		\centering
		\subfloat[\label{r1gt}]{\includegraphics[width=0.475\columnwidth]{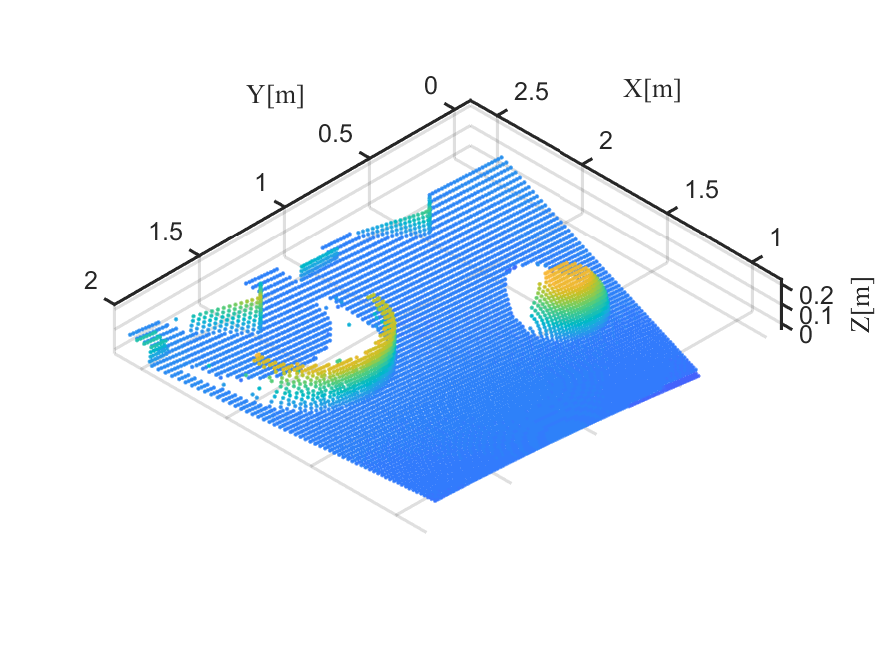}}\enskip 
		\subfloat[ \label{r1s}]{\includegraphics[width=0.475\columnwidth]{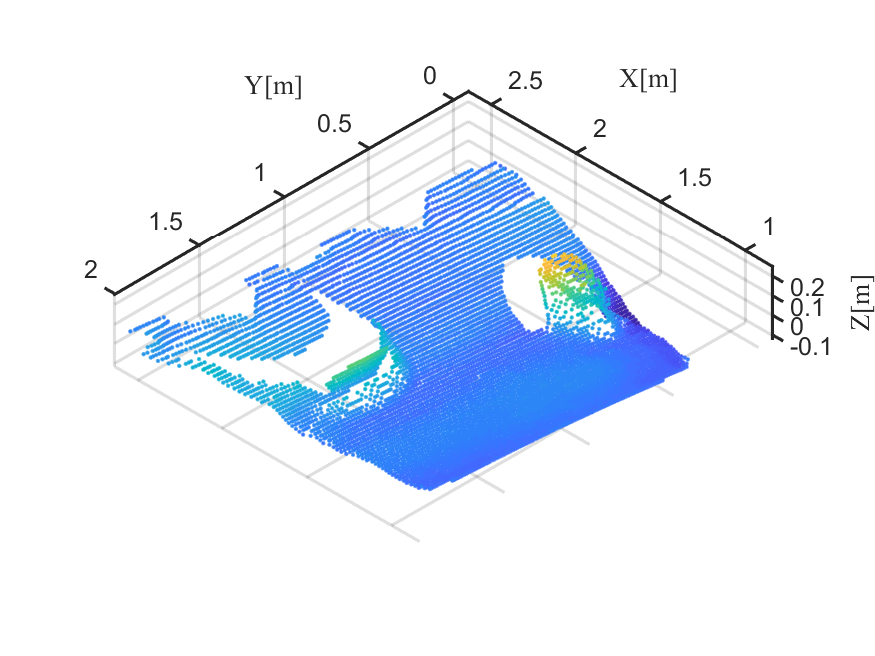}}\\
            \subfloat[\label{r1usm}]{\includegraphics[width=0.475\columnwidth]{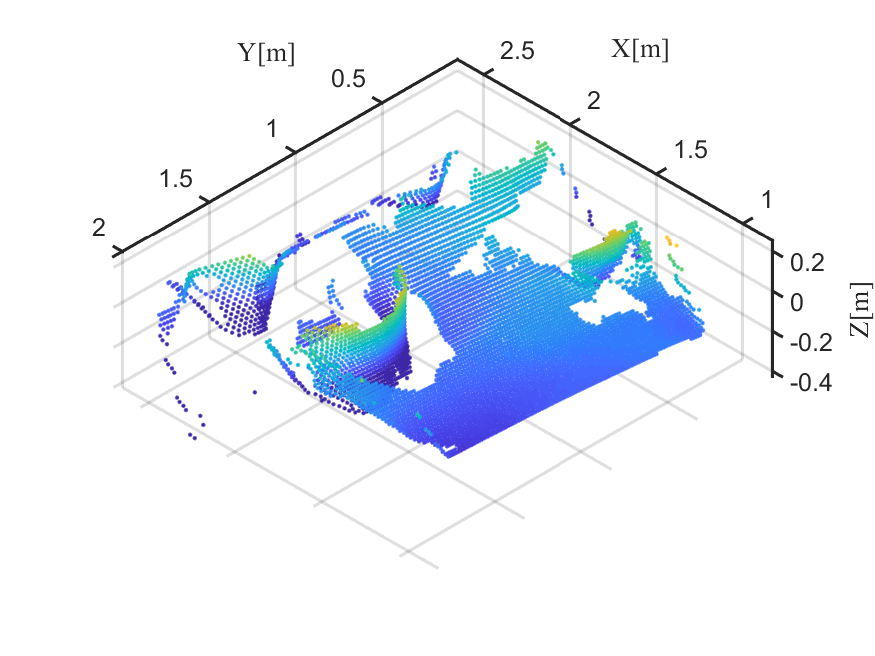}}\enskip 
		\subfloat[ \label{r1us}]{\includegraphics[width=0.475\columnwidth]{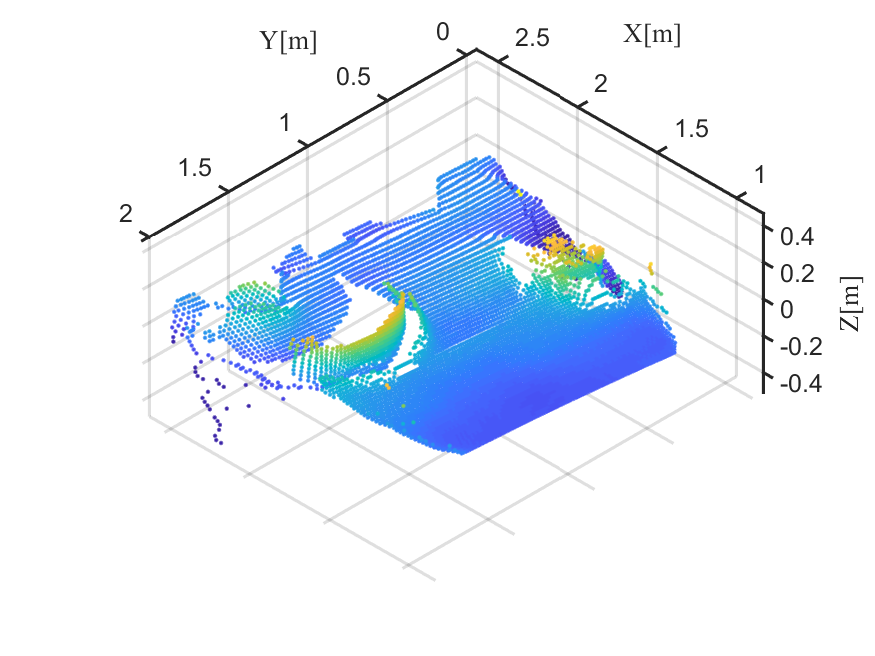}}
		\caption{Estimation results of Fig.~\ref{fig:img} in the point cloud: (a) the ground truth, (b) the result of supervised learning, (c) the result without using the learned signal mask, the mask is from simple binarization with a threshold, and (d) the result of the proposed method. }  
		\label{fig:real_result}
	\end{center}
\end{figure}

Figure~\ref{fig:real_result} presents the estimation results of Fig.~\ref{fig:img} in terms of point cloud. As shown in Fig.~\ref{fig:real_result}(d), the proposed method successfully estimated the 3D structure of the object. The result of the arc in Fig.~\ref{fig:real_result}(d) is better than that shown in Fig.~\ref{fig:real_result}(b) from supervised learning. There will inevitably be some distortions in the point cloud from self-supervised learning because there will always be regions that do not overlap between the consecutive frames. The result in Fig.~\ref{fig:real_result}(c) is noisier because of the mask; however, the 3D structure can still be successfully learned. 


\section{Conclusion}
In this study, we dug deep into the self-supervised learning of the elevation angle estimation problem for FLS. It is found that many failure cases may be caused by the motion degeneracy problem. We analyzed the motion field and identified efficient motions for network training. We proved that for basic motions, $t_z$ and $\omega_x$ motions can contribute to the learning process. We also showed that pretraining using synthetic data is not a must if the dataset is properly built. The motion field analysis conclusions may not only contribute to self-supervised learning through inverse warping signal but may also be important to simultaneous localization and mapping and multi-view 3D reconstruction problems. 

Future work may include auto-filtering degenerate motions from a dataset with various motions. Signal masks should be further studied. In this study, the motion between two consecutive frames was considered given. It may also be worth discussing whether motions can be learned in a self-supervised manner simultaneously.  
\printbibliography

\addtolength{\textheight}{-12cm}   




\end{document}